\documentclass[sigconf]{acmart}
\AtBeginDocument{%
  }

\setcopyright{acmlicensed}
\copyrightyear{2025}
\acmYear{2025}
\acmDOI{10.1145/3746027.3755116}
\acmConference[MM '25]{Proceedings of the 33rd ACM International Conference on Multimedia}{}{October 27--31, 2025,Dublin, Ireland.}
\acmISBN{979-8-4007-2035-2/2025/10}

\begin{document}

\title{Dual-Learning based Penalized Multi-Align Clustering for Multi-View Incomplete and Disorderly Data}

\author{Liang Zhao}
\orcid{0000-0001-6301-1311}
\affiliation{%
  \institution{Dalian University of Technology}
  \city{Dalian}
  \state{Liaoning}
  \country{China}
}
\email{liangzhao@dlut.edu.cn}

\author{Shubin Ma}
\affiliation{%
  \institution{Dalian University of Technology}
  \city{Dalian}
  \state{Liaoning}
  \country{China}}
\email{1369830844@mail.dlut.edu.cn}

\author{Bo Xu}
\authornote{Corresponding author.}
\affiliation{%
  \institution{Dalian University of Technology}
  \city{Dalian}
  \state{Liaoning}
  \country{China}}
\email{boxu@dlut.edu.cn}

\author{Qingchen Zhang}
\affiliation{%
  \institution{Hainan University}
  \city{Haikou}
  \state{Hainan}
  \country{China}}
\email{zhangqingchen@hainanu.edu.cn}

\renewcommand {\shortauthors}{Liang Zhao, Shubin Ma, Bo Xu,  Qingchen Zhang}


\begin{abstract}
Multimodal feature fusion, by integrating the complementary information from each modality, can effectively capture complex features in real-world data. However, in many use cases, such as boiler combustion monitoring, factors including equipment failure, inconsistent sensor sampling frequencies, and network delays often cause data collected from different modalities to suffer from missing modality and temporal asynchrony. This leads to the incompleteness and disorderliness of multimodal data. To address these issues, previous studies have proposed several data fusion methods that align the cluster centers before fusion. However, these approaches have two key limitations: 1) they do not guarantee a high alignment accuracy of data pairs at the sample level, and 2) they do not address the issue of significant discrepancies in data sizes across different classes, which impacts the subsequent data fusion performance.

To these ends, we propose a dual-learning based penalized multi-align clustering model (DLPMAC). The Dual-Learning mechanism in the model ensures that it can learn the prior knowledge inherent in each modality’s data, specifically the semantic and structural information. This facilitates the maintenance of both semantic consistency and structural similarity—on both local and global levels—across different modalities. Additionally, the Penalized Multi-Align module achieves multi-to-multi data alignment based on a penalty mechanism, which enables a single sample to form data pairs with different samples from other modalities, thereby enhancing the alignment accuracy of data pairs. The introduction of the penalty mechanism prevents data aggregation phenomena, thereby avoiding situations where excessive samples are linked to a single sample. Experimental results validate the efficacy of this model in addressing data alignment(from the sampling and the clustering perspectives) and fusion challenges. The code and related supporting materials are available at: \url{https://github.com/Autism-mm/DLPMAC}.
\end{abstract}


%


\maketitle

\section{Introduction}
\label{sec:Introduction}

The multimodal clustering\cite{zhao2023multi,zhao2024learnable,zhou2024survey,ma2025consistency,zhang2016multi,dong2023cross,chen2023deep} method, by utilizing multimodal data\cite{wang2022am3net, lian2023gcnet, zong2024self, zhu2024vision+}, can provide more comprehensive and rich information than a single modality. In the application of multimodal data, the alignment of complete data across different modalities offers complementary information for data fusion\cite{zhao2024deep}. However, in industrial data collection, multimodal data (such as pressure, temperature, images, etc.) often suffer from missing values and misalignment due to equipment malfunctions, inconsistent sensor sampling frequencies, environmental factors, and other problems. As a result, the collected multimodal exhibit missing values and also suffer from misalignment\cite{yu2021novel,zong2018multi}, meaning that the number of data points across different modalities is inconsistent and disordered. Incompleteness\cite{yuan2024robust,du2024fast,wen2023scalable,zhao2023unrestricted} refers to the absence of certain data in some modalities, while misalignment refers to the inability to establish a one-to-one correspondence between data items across different modalities.

\begin{figure}
    \centering
    \includegraphics[width=0.9\linewidth]{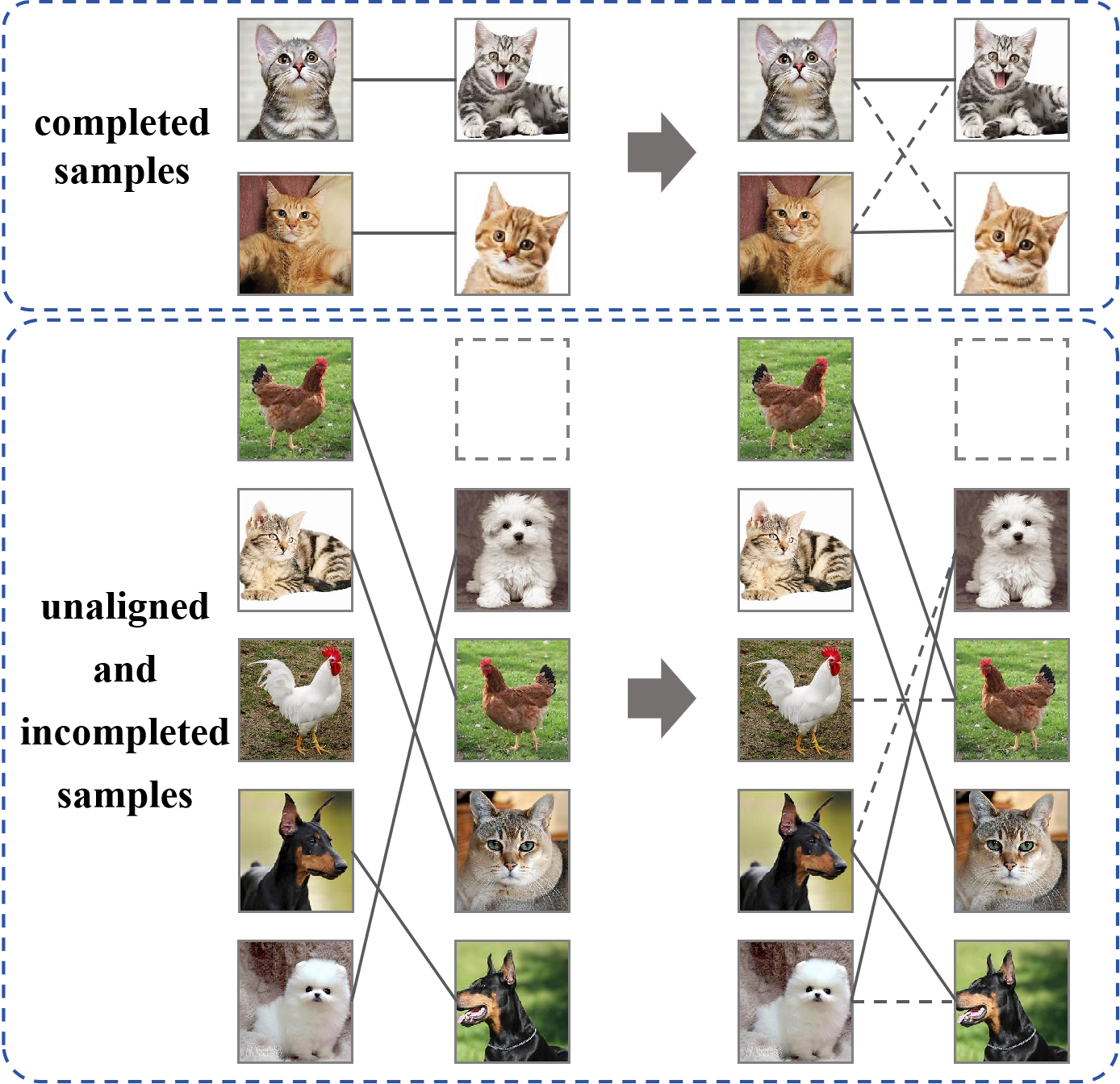}
    \caption{Example graph of incomplete and misaligned data}
    \Description{
The figure illustrates two sets of image samples from different modalities. The top section shows fully aligned and completed samples with consistent pairings across modalities. The bottom section shows unaligned and incomplete samples, with some images missing and connections between images mismatched or inconsistent. Arrows indicate the transformation from disordered data to aligned structures. The images include cats, dogs, chickens, and other animals, demonstrating multimodal data misalignment and incompleteness.
}
    \label{fig:ipaeg}
\end{figure}

Figure \ref{fig:ipaeg} provides a toy example of clustering\cite{mao2023congmc,zhang2024unsupervised,guo2024robust,li2023orthogonal,dai2024multi,abbas2023clusterfug} incomplete and misaligned\cite{ren2024novel} multimodal data. Solid lines indicate that two data points belong to the same pair, while dashed lines indicate that they belong to the same class. The multimodal data are divided into two parts: the fully aligned multimodal data and the incomplete and misaligned multimodal data. In the second part, the data from the two modalities are not only inconsistent in quantity but also lack corresponding relationships across modalities. For example, in road traffic monitoring research, the system uses multiple images to determine whether a vehicle is violating regulations or speeding. However, the images of the same vehicle are not captured consecutively. Between the capture of two images, the camera continues to capture images of other vehicles, and there may be missed shots due to vehicle occlusion or system issues. This leads to the phenomenon of incomplete and misaligned multimodal data in the application.

The main challenge in dealing with incomplete and disordered multimodal data\cite{deng2025reproducibility} is how to align and fuse the incomplete and disordered data effectively. Due to the data being disordered and lacking label information\cite{chen2024partial}, it becomes particularly difficult to leverage both intra-modal and inter-modal information for data completion. Currently, research on the issues of incomplete and misaligned multimodal data is relatively scarce, with previous work mainly focusing on addressing data misalignment. In tackling this issue, researchers have optimized the process in two key areas. First, by preprocessing the initial data to enhance its quality and consistency. For example, Qian et al.\cite{qian2024nonparametric} proposed a deep learning-based variational inference framework (DeepVNC) that automatically extracts latent clustering information from data without strict labels, uncovering underlying structures and guiding cross-view learning. He et al.\cite{he2024robust} introduced the VITAL framework, which combines variational inference with contrastive learning, modeling data samples as Gaussian distributions in latent space, where the mean represents shared information and the variance captures view-specific details. Second, researchers have improved network design and training to strengthen the model's feature extraction capabilities. For instance, Yang et al.\cite{yang2021partially} designed a noise-resistant contrastive loss to reduce the impact of false-negative pairs. Zeng et al.\cite{zeng2023semantic} integrated autoencoders with cross-entropy losses to reduce cross-modal differences and improve semantic distinctions. Zhao et al.\cite{zhao2024dynamic} applied GCN for identifying difficult-to-align representations and dynamically optimized the adjacency graph, progressively learning consistency across multiple views for more reliable alignment. In addressing the issues of incompleteness and misalignment, Yang et al.\cite{yang2022robust} directly handle zeroed incomplete and misaligned multimodal data, using the Hungarian algorithm for alignment and filling in missing data(SURE). Zhao et al.\cite{zhao2024incomplete}, on the other hand, align data at the class level and then perform feature fusion across multimodal classes without the need to fill in missing instances or sequence information. However, both approaches fail to ensure high alignment accuracy at the sample level and do not effectively address the issue of significant data size differences between different classes, which subsequently affects the data fusion process.

To ensure high alignment accuracy of data pairs at the sample level, we propose a Dual-Learning mechanism that enables the model to learn prior knowledge, such as semantic consistency within multimodal classes, correlations between data, and similarities in both local and global structures of multimodal data. Combined with contrastive learning, the model further enhances its understanding of the semantic consistency within classes and the semantic distinction between classes. To address the issue of significant differences in data quantities across classes, we design a Penalized Multi-Align module. This module facilitates multi-to-multi alignment of multimodal incomplete and disordered data, allowing a single sample to form data pairs with multiple samples from other modalities for data fusion. Moreover, the introduction of the penalty mechanism effectively prevents clustering phenomena (i.e., one sample aligning with too many others). Experimental results demonstrate that this approach is significantly effective in solving the issues of multimodal data incompleteness and misalignment.

The contributions of this paper can be summarized below.

\begin{itemize}

\item We are the first to study the alignment problem from the sampling perspective, addressing data alignment from the sampling perspective and the clustering perspective. Experiments carried out on \textbf{16} datasets demonstrate that the model shows strong versatility and effectiveness in sample-level alignment tasks.

\item We propose a Dual-Learning mechanism that learns both semantic and structural information across multimodal data, fully considering the semantic consistency within classes and the local and global structural similarities between different modalities.

\item We design a Penalized Multi-Align module that resolves the issue of significant data size differences between classes through multi-to-multi alignment. The introduction of the penalty mechanism effectively prevents clustering phenomena during the data alignment process.
 
\end{itemize}

\section{The Proposed Method}
\label{sec:The Proposed Method}

Given a dataset 
\(X=\{X^{(1)},X^{(2)},X^{(3)},\dots,X^{(v)}\}\) ,\({X}^{(v)} \in {R}^{n \times dv}\), represents the data of the \emph{vth} modality with \emph{n} samples and dimensionality \emph{dv}. \(X=\{A,B\}\), where \(A \in R^{n_a \times dv}\) represents the fully aligned data and \(B \in R^{n_{b} \times dv}\) represents the incomplete and misaligned data. The data in \emph{B} is disordered, and the missing values in \emph{B} are handled based on the incompleteness rate \(\eta\).

To address the issue of low alignment accuracy and poor data fusion performance in incomplete and disordered multimodal data, we propose DLPMAC. In the model, we effectively learn multimodal data's semantic and structural information through sample-level Dual-Learning. This ensures the consistency of similar data information across modalities and the similarity of modal structures. Additionally, by utilizing the Penalized Multi-Align module, we efficiently leverage the information from incomplete and disordered data, thereby solving the problems of low alignment accuracy and poor data fusion. The specific methodology is as follows.

\begin{figure*}
	\centering
		\includegraphics[scale=0.54]{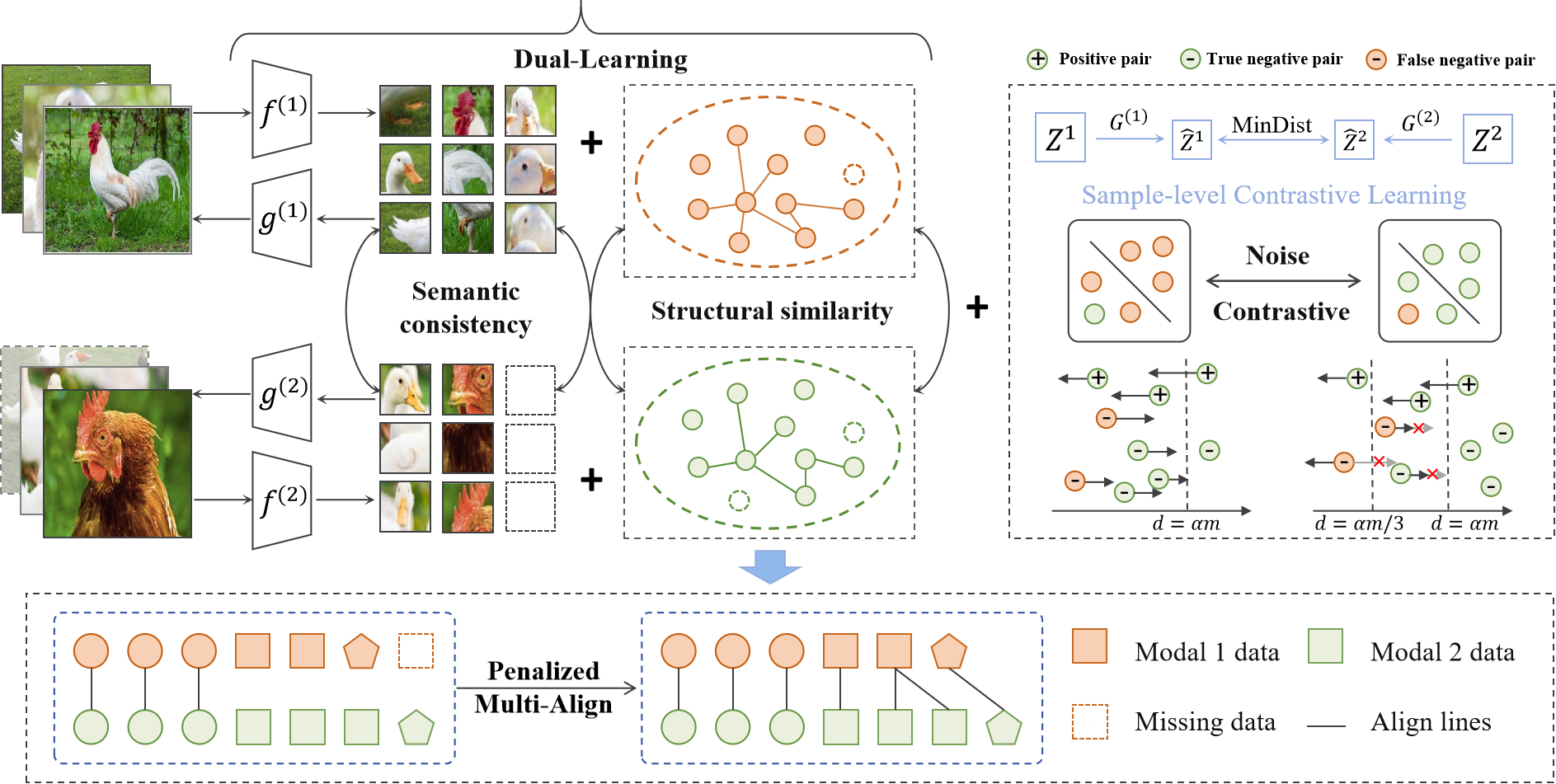}
	\caption{Incomplete and unaligned DLPMAC model. In the Dual-Learning mechanism, latent features are learned through semantic consistency of data and structural similarity between modalities, while noise-contrastive learning removes some false negative pairs and further distinguishes data from different classes. In the Penalized Multi-Align module, a multi-to-multi approach is used to align incomplete and disordered data streams from different modalities, effectively solving the issues of incompleteness and misalignment in multimodal data.}
    \Description{ An overview of the DLPMAC model, which integrates Dual-Learning and Penalized Multi-Align modules. The upper part shows Dual-Learning composed of semantic consistency, structural similarity, and sample-level noise-contrastive learning. The lower part illustrates the Penalized Multi-Align module that aligns incomplete and disordered data streams from two modalities. Modal 1 and Modal 2 samples are represented by various shapes and colors, with dashed outlines indicating missing data and lines showing alignment relations.}
	\label{model}
\end{figure*}

\subsection{Dual-Learning}
Data belonging to the same class across different modalities exhibit high semantic consistency, whereas data from different classes demonstrate significant semantic divergence. Based on this, by training with complete cross-modal aligned data, the model can effectively capture the semantic correlations between data from different modalities, thus learning the semantic consistency among data of the same class across modalities. The loss function is as follows

\begin{equation}
    L_1 = \left( \sum_{i\neq j}^{m}\sum_{k=1}^{n} \left(f_i(a_k^i) - f_j(a_k^j)\right)^2 \right)^{\frac{1}{2}} .
\end{equation}

To ensure that the model can accurately capture the latent features of the data, we introduce a reconstruction loss to enforce semantic consistency between the data before and after training

\begin{equation}
    L_2 = \left( \sum_{i=1}^{m}\sum_{k=1}^{n} \left(a_k^i - d_i\left(f_i(a_k^i)\right)\right)^2 \right)^{\frac{1}{2}} ,
\end{equation}

where, \(a_k^i\) represents the \emph{kth} data of the \emph{ith} modality,  
\(f_i(\cdot)\) denotes the encoder of the \emph{ith} modality ,and \(d_i(\cdot)\) denotes the decoder of the \emph{ith} modality. After the model has initially learned the information associations between the data, it jointly learns the structure between modalities based on the semantic information of the data. During each training step, a small portion \(A_{b}\) is extracted from the fully aligned data 
\(A\), and then scrambled using a permutation matrix \(P_a\) to obtain \(\bar{A}_b\).For the \(i\), \(j\) modalities 

\begin{equation}
\begin{cases}
f_i(\bar{A_{b}^i})=f_i(A_{b}^i) \\
f_j(\bar{A_{b}^j})=P_a \cdot f_j(A_{b}^j)
\end{cases}
.
\end{equation}

To re-align the data of the two modalities, the model uses the Hungarian algorithm to learn a permutation matrix \(\bar{P}_a\)

\begin{equation}
L_3=\left\|\bar{P_a}-P_a\right\|_{F} .
\end{equation}

Through iterative optimization, the model progressively learns the local structural features of the data while maintaining local structural consistency across different modalities. As the number of training epochs increases, the model effectively accumulates local structural information, enabling precise modeling of the global structure across modalities.

To further aggregate data of the same class across modalities and distinguish data of different classes, we introduce a cross-modal contrastive loss function

\begin{align}
\label{contras}
L_4 &= \frac{1}{2n} \sum_{i=1}^{n} \left( Y \cdot {D(z_k^i,z_h^j)} + (1 - Y) \cdot \frac{1}{m} \right. \\
&\quad \left. \max \left( \alpha \cdot m \cdot {D(z_k^i,z_h^j)}^{\frac{1}{2}} - {D(z_k^i,z_h^j)}^{\frac{3}{2}} , 0 \right)^2 \right) \notag ,
\end{align}

where \(z=f(x)\) represents the latent feature representation of the data, \textbf{Y} represents the labels of positive and negative data pairs, \(D(x_k,x_h)=\left\|f_i(x_k^i)-f_j(x_h^j)\right\|_2^2\), and \(\alpha\) is a hyperparameter that controls the distance range within which false negative pairs can be mitigated. 

The final loss function of Dual-Learning is as follows

\begin{equation}
L_{dl}=L_1+L_2+\lambda L_3+L_4 .
\end{equation}

Here, \(\lambda\) is set to 1 and can be fine-tuned. Within the Dual-Learning mechanism, the model leverages the supervision from aligned data to enhance the accuracy of cross-modal retrieval.

\subsection{Penalized Multi-Align}

\emph{Multi-to-multi sample-level alignment mechanism:}
During the cross-modal alignment process, we ignore the missing data within each modality and transform the problem into an asymmetric multimodal alignment task, where the data samples across modalities are unequal in number. The incomplete and disordered data are collectively represented as the set \emph{B}.

\begin{figure}[H]
    \centering
    \includegraphics[width=0.9\linewidth]{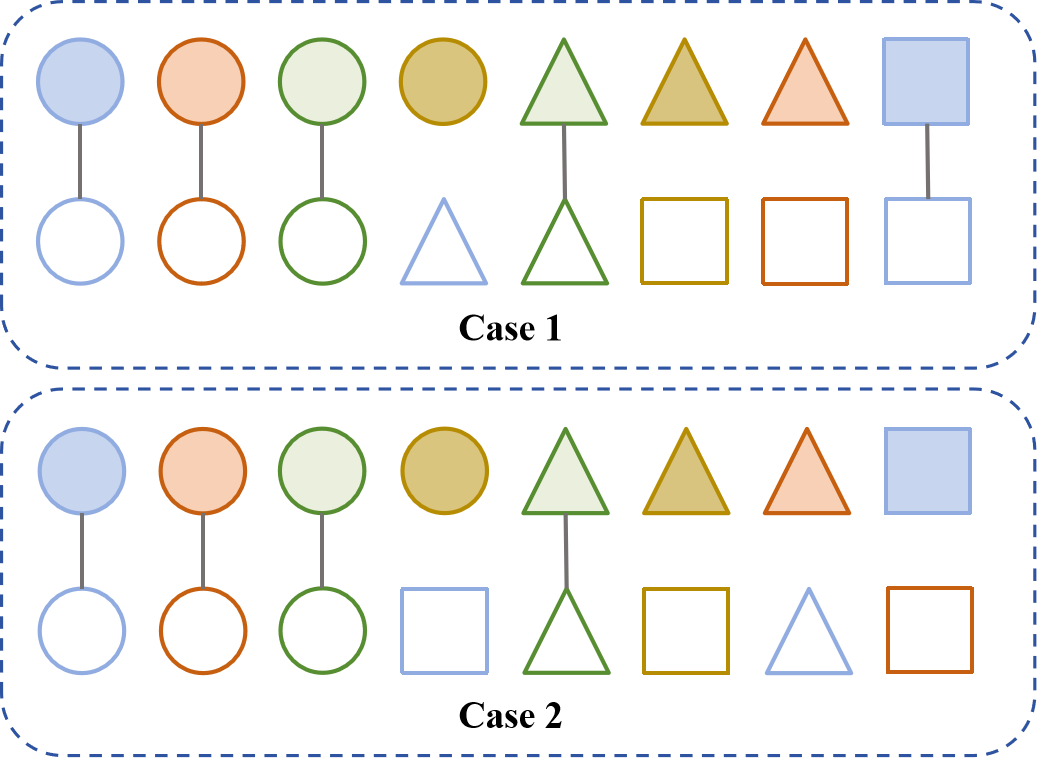}
    \caption{Pairwise alignment of data with varying numbers within the same class.}
    \Description{
Two alignment cases showing pairwise data matching between two modalities with different sample counts per class. In each case, colored shapes represent samples of different classes, and lines connect paired samples across modalities. Case 1 shows partial alignments where some samples are unmatched due to number imbalance. Case 2 illustrates an alternative alignment for the same classes. Shapes include circles, triangles, and squares in various colors, indicating distinct classes.
}
    \label{fig:incomalign}
\end{figure}

Due to the incompleteness and misalignment of multimodal data, the order of some data may become disrupted, and the distribution of data across classes may be uneven. This characteristic can lead to errors in the alignment process between different modalities, causing data from different classes to be incorrectly matched into pairs, as shown in Figure \ref{fig:incomalign}. In Figure \ref{fig:incomalign}, each row represents a different modality, with markers of the same shape indicating the same class, while markers with the same shape and color represent correctly aligned data pairs.

\textbf{First case:} When there is a significant disparity in the number of data points across different classes, the class with a larger amount of data will dominate the alignment process, forcing the data from other classes to be incorrectly matched with it. This misalignment issue propagates layer by layer, ultimately leading to a significant decrease in the proportion of correctly aligned data in the smaller classes. As shown in Figure \ref{fig:incomalign}, in this case, the alignment accuracy can only reach 50\%.

\textbf{Second case:} Suppose that data \(B_k^1\) and \(B_k^2\) belong to the same semantic class, but due to the highest similarity between \(B_h^1\) and \(B_k^2\) across modalities, they are incorrectly matched as an aligned pair. This misalignment causes the \(B_k^1\) data in modality \emph{k} (represented as a blue square in Figure \ref{fig:incomalign}) to deviate from its true correspondence, further reducing the alignment accuracy. To address the aforementioned issues, a sample-level alignment and class-level fusion strategy is employed. The specific computation formula is as follows

\begin{equation}
    s_{kh}^{ij}=
\begin{cases}
 & \arg\max S(b_{k}^{i},b_{h}^{j}), {h\in[1,n_{b}]} \quad\mathrm{if}\quad S(b_{k}^{i},b_{h}^{j})>0 \\
 & 0,\quad\mathrm{if}\quad no \ match \ exists \ for \ b_{h}^{j}  
\end{cases}
.
\end{equation}

Here, \(s_{kh}^{ij}\)  represents the similarity between the 
\emph{kth} data of the \emph{ith} modality and the \emph{hth} data of the 
\emph{jth} modality. \emph{S} denotes the alignment matrix, which stores the alignment information and similarity between the data in the two modalities.

\begin{figure}
    \centering
    \includegraphics[width=0.9\linewidth]{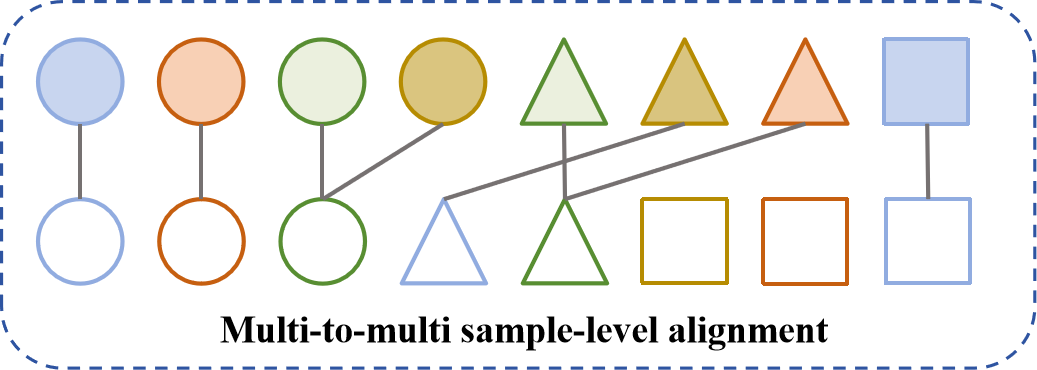}
    \caption{Multi-to-multi alignment results}
    \Description{An illustration of multi-to-multi sample-level alignment. The figure contains different shapes including circles, triangles, and squares in various colors, representing samples from different modalities. Lines connect multiple shapes across modalities, indicating sample-level alignment relations beyond one-to-one mapping.}
    \label{fig:multi-to-multi-alignment}
    \label{fig:multi}
\end{figure}

As shown in Figure \ref{fig:multi}, a data point can be aligned with multiple other data points; however, it is also possible that a data point does not align with any other data, as indicated by the two dashed rectangles. Finally, after obtaining the alignment relationships between different modalities, the latent feature representations from different modalities are concatenated and fused based on these alignment relationships, generating a unified multimodal clustering representation.

\begin{figure}
    \centering
    \includegraphics[width=1\linewidth]{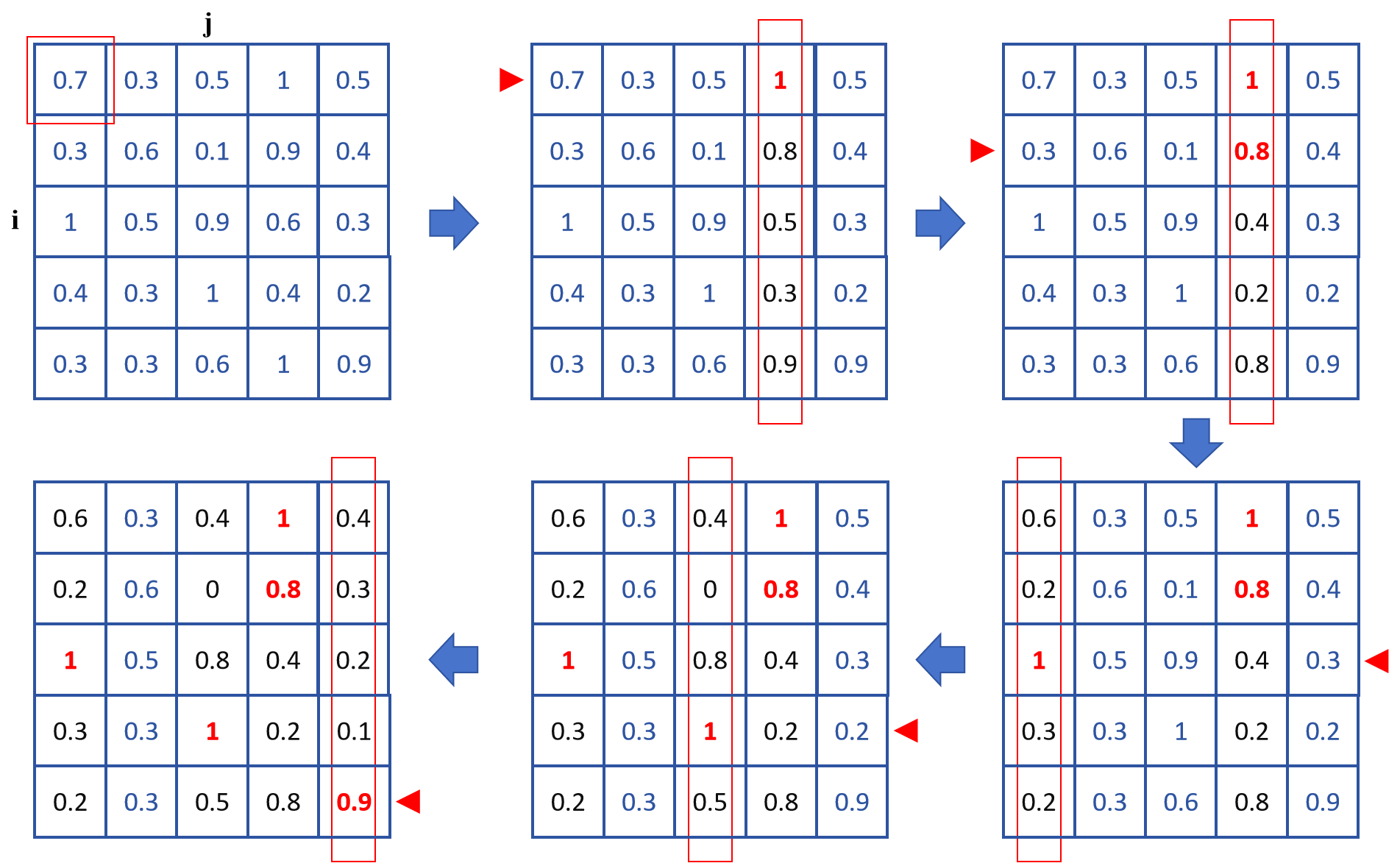}
    \caption{Red and bold values indicate the similarity scores between aligned data pairs. Each time a pair is marked, their similarity is reduced by 0.1.}
    \Description{Six matrices showing similarity scores between data pairs. Red bold numbers represent aligned pairs, whose similarity decreases by 0.1 each time they are marked. Arrows indicate the iterative process of similarity reduction across different matrices.}
    \label{fig:RAB}
\end{figure}

In the multi-to-multi alignment process, it is possible for a single data point to align with too many other data points, leading to data aggregation. To effectively mitigate this phenomenon, we introduce a penalty mechanism during the alignment process. This is achieved by dynamically reducing the similarity between already aligned data points to optimize the alignment, as illustrated in Figure \ref{fig:RAB}.

As seen in Figure \ref{fig:RAB}, the \emph{4th} data point of the \emph{jth} modality is bidirectionally aligned with both the \emph{1st} and \emph{2nd} data points of the \emph{ith} modality, while the \emph{2nd} data point of the \emph{jth} modality does not exhibit any valid alignment relationships.

\begin{table*}[!htbp]
\centering
\caption{The \(acc_s\) of seven different algorithms across sixteen datasets}
\label{tblsa}
\resizebox{\textwidth}{!}{
\begin{tabular}{l|cccccccc}
    \hline
    \multicolumn{8}{c}{Datasets}\\
    \hline
    \textbf{Methods} & \textbf{3Sources} & \textbf{BBCsports} & \textbf{Caltech101} & \textbf{BDGP}&\textbf{HandWritten}&\textbf{Movies}&\textbf{flower17}&\textbf{Prokaryotic} \\
    \hline
    PVC(20'NeurIPS)    & \underline{0.4688} & \underline{0.3784} & 0.4641 & 0.7880 &\underline{0.7930} &0.1459&\underline{0.3037}&0.4207 \\
    MvCLN(21'CVPR)  & 0.3254 & 0.3546 &0.2624  &0.2676  &0.0955  &0.0600&0.0570&0.4111\\
    SURE(22'TPAMI)   & 0.2249 & 0.3014 & 0.2309 &0.2572 & 0.0910&0.0843&0.0559&0.3230 \\
    SMILE(23'TPAMI)  & 0.3852 & 0.3443 & 0.3252 &0.3448  & 0.1735 & 0.0665&0.0934&\underline{0.6339}\\
    EGPVC(23'ICASSP)  & 0.4178 & 0.3598 & \underline{0.4803} & 0.7916 &0.6685 &0.1540&0.2882&0.3964 \\
    DGPPVC(24'TNNLS)  & 0.3964 & 0.3546 &0.3550  & \underline{0.8144} & 0.4995 &\underline{0.1669}&0.0588&0.3848\\
    DLPMAC  &\textbf{0.9379} & \textbf{0.7507} & \textbf{0.8728} & \textbf{0.9240} & \textbf{0.8260} &\textbf{0.6335}&\textbf{0.4589}&\textbf{0.8404}\\
    \hline
    \textbf{Methods} & \textbf{yale\_mtv}	&\textbf{Reuters\_dim10}&	\textbf{ORL} & \textbf{MSRCv1}&	\textbf{20NewsGroups}&	\textbf{BBC4}&	\textbf{ALOI}	&\textbf{Wikipedia-test} \\
    \hline
    PVC(20'NeurIPS)    & \underline{0.4661} & 0.4416 & 0.355 & \underline{0.7333}& 0.2980&0.3401&0.2827&\underline{0.3723} \\
    MvCLN(21'CVPR)  & 0.0636 &0.2344  &0.0337  &0.15  & 0.2320 &0.2781&0.0132&0.1400\\
    SURE(22'TPAMI)   &  0.0788 & 0.2709 &  0.0525&0.2238 &0.2040 &0.2131&0.0180&0.0967 \\
    SMILE(23'TPAMI)  & 0.1594 & \underline{0.5093} & 0.1400 & 0.2048 &0.2340  &0.2628&0.0980&0.1313\\
    EGPVC(23'ICASSP)  & 0.4333 & 0.4397 & \underline{0.4225}&  0.7286 &0.2740&0.4029&\underline{0.315}8&0.3709 \\
    DGPPVC(24'TNNLS)  & 0.1212 &  0.3983& 0.1000 & 0.1667 & \underline{0.3360} &\underline{0.5401}&0.3154&0.3045\\
    DLPMAC  &\textbf{0.8490}  & \textbf{ 0.7193}& \textbf{0.7695} & \textbf{0.9261} & \textbf{0.7653} &\textbf{0.8772}&\textbf{0.7622}&\textbf{0.6501}\\
\hline
\end{tabular}}
\end{table*}

\begin{table*}[!htbp]
    
    \centering
    \caption{Clustering performance (ACC, NMI, F1) on datasets with alignment rate 0.5 and incomplete rate 0.5}
    \label{tbl0.5}
    \resizebox{\textwidth}{!}{
    \begin{tabular}{cc|ccccccc}
        \hline
            & \multicolumn{7}{c}{Methods} \\
        \hline
        Datasets &  & PVC 20' & MvCLN 21' & EGPVC 22' & SMILE 23'& SURE 23'& DGPPVC 24'& Ours \\
        \hline
            &ACC& 0.4686 & 0.3645 & 0.4278 & 0.3462 & 0.2823 & \underline{0.4905}&\textbf{0.6346} \\
        3Sources&NMI&\underline{0.3354} & 0.1039 & 0.2182 & 0.0918 & 0.1256 & 0.2270&\textbf{0.5423}\\
            &F1& \underline{0.4836} & 0.2809 & 0.4331 & 0.3091 & 0.3515 & 0.4048&\textbf{0.6568} \\
    
        \hline
            &ACC & 0.7880 &0.3396 & 0.7916 & 0.4644 & 0.5314 &   \underline{0.8144} &\textbf{0.9403}\\
            BDGP&NMI & 0.5720 &0.1060 & 0.5549 & 0.1658 & 0.4151 &  \underline{0.5767}&\textbf{0.8372} \\
            &F1& 0.7899&  0.3137 &0.7933 & 0.4545 & 0.5426 &  \underline{0.8109}&\textbf{0.9407} \\
    
       \hline
            
            &ACC& \underline{0.4416} &0.3211  & 0.4397 & 0.4304 & 0.3202 &  0.3983&\textbf{0.5257}\\
        Reuters\_dim10&NMI&0.2090 & 0.04 &0.1894 & \underline{0.2099} & 0.0692  &0.0764& \textbf{0.3173} \\
             &F1& 0.4410& 0.2278 & 0.4219 & \underline{0.4603} & 0.4322 & 0.3452&\textbf{0.5531} \\

        \hline
            &ACC&\underline{0.7333} & 0.3929 & 0.7289 & 0.4762 & 0.5166 &  0.1667 &\textbf{0.7748}\\
            MSRCv1&NMI & \underline{0.6601} & 0.2313 & 0.6035& 0.3380 & 0.4188   & 0.0651&\textbf{0.6694} \\
             &F1&\underline{0.7334} & 0.3924 & 0.7201 & 0.4534 & 0.5213 &   0.0802&\textbf{0.7744} \\
        \hline
            &ACC & 0.2827 & 0.1857 &0.3158 & 0.2405 & \underline{0.3236} &  0.3154&\textbf{0.7744}\\
      ALOI &NMI & 0.5382 & 0.3542 & 0.5556 & 0.4468 & \underline{0.5719} &  0.4772 & \textbf{0.8488}\\
             &F1& 0.2598&0.1790 & 0.2936 &0.2247 & \underline{0.4013} & 0.3099&\textbf{0.7730} 
            \\
              \hline
        &ACC & 0.4333 &0.3437 & \underline{0.4588} & 0.4109 & 0.3721 &  0.2751 &\textbf{0.5861}\\
        yale\_mtv&NMI & 0.5182 & 0.4044 & \underline{0.5335} & 0.4796 & 0.4529 &  0.3749&\textbf{0.6135} \\
        &F1& 0.4189 & 0.3288 & \underline{0.4465} & 0.4090 & 0.3611  & 0.2198&\textbf{0.6070} \\
    
       \hline
      
    \end{tabular} }
\end{table*}

\begin{table*}[!htbp]
   
    \centering
    \caption{Clustering performance (ACC, NMI, F1) on datasets with alignment rate 0.3 and incomplete rate 0.5}
    \label{tbl0.3}
    \resizebox{\textwidth}{!}{
    \begin{tabular}{cc|ccccccc}
        \hline
            & \multicolumn{7}{c}{Methods} \\
        \hline
        Datasets &  & PVC 20' & MvCLN 21' & EGPVC 22' & SMILE 23'& SURE 23'&DGPPVC 24'& Ours \\
        \hline
            &ACC&  0.3710& 0.3538 &0.4414 & 0.3643 & 0.3426   &\underline{0.4651} &\textbf{0.6548} \\
        3Sources&NMI&0.2210&0.0996&0.2247&0.1122&0.1834&\underline{0.2386}&\textbf{0.4935}\\
            &F1& 0.3757&0.2794&\underline{0.4286}&0.3222&0.3971&0.3747&\textbf{0.6434} \\
    
        \hline
            &ACC & \underline{0.7596} &0.3300 & 0.6336 & 0.3802 & 0.4115 &   0.6384 &\textbf{0.9039}\\
            BDGP&NMI & \underline{0.5157} &0.0988 & 0.3784 & 0.1243 & 0.2488 & 0.3703 &\textbf{0.7635} \\
            &F1&\underline{0.7631} &  0.2940& 0.6483 & 0.3761 & 0.4334 &  0.6340&\textbf{0.9040} \\
    
       \hline
            
            &ACC& 0.4447 & 0.3394 & 0.4277 & \underline{0.4468} & 0.4016 &  0.2737&\textbf{0.5319}\\
        Reuters\_dim10&NMI& 0.2136&  0.0825& 0.2145& \underline{0.2240} & 0.1773 &  0.0039& \textbf{0.3053} \\
             &F1&0.4300 & 0.3093 & 0.4046 & 0.4738 & \underline{0.5271} & 0.1196&\textbf{0.5512} \\

        \hline
            &ACC& \underline{0.7381}& 0.4222 & 0.6381 & 0.4313 & 0.5450 &   0.3762&\textbf{0.7669}\\
            MSRCv1&NMI & \underline{0.6182} & 0.2604 &0.5156 & 0.2829 & 0.4552 &  0.3095 &\textbf{0.6255} \\
             &F1& \underline{0.7358}& 0.4217 & 0.6353 & 0.4097 & 0.5498 &   0.2852&\textbf{0.7670} \\
        \hline
            &ACC & 0.2976 &0.1886  &0.2819 & 0.2617 & 0.3080 &  \underline{0.3779}&\textbf{0.7226}\\
      ALOI &NMI & \underline{0.5624} & 0.3448 & 0.5364 & 0.4520 & 0.5403 &  0.5295 & \textbf{0.8202}\\
             &F1&0.2571 &0.1929 & 0.2603 & 0.2457& \underline{0.4117} & 0.3715&\textbf{0.7181} 
            \\
              \hline
        &ACC & 0.4364&0.3297&\underline{0.4424}&0.4388&0.3636&0.1830&\textbf{0.5038}\\
        yale\_mtv&NMI &0.5036&0.4106&\underline{0.5228}&0.4995&0.4361&0.2480&\textbf{0.5439} \\
        &F1&\underline{0.4339} &0.3080&0.4270&0.4336&0.4103&0.1640&\textbf{0.5370}\\
    
       \hline
      
    \end{tabular} }
\end{table*}

\begin{table*}[!htbp]
    \centering
    \caption{Clustering performance (ACC, NMI, F1) on datasets with alignment rate 0.7 and incomplete rate 0.5}
    \label{tbl0.7}
    \resizebox{\textwidth}{!}{
    \begin{tabular}{cc|ccccccc}
        \hline
            & \multicolumn{7}{c}{Methods} \\
        \hline
        Datasets &  & PVC 20' & MvCLN 21' & EGPVC 22' & SMILE 23'& SURE 23'& DGPPVC 24'& Ours \\
        \hline
            &ACC& 0.4414&0.3539&0.4201&0.3118&0.3698&\underline{0.4716}&\textbf{0.6362} \\
        3Sources&NMI&\underline{0.3303}&0.0874&0.2183&0.0863&0.2116&0.2445&\textbf{0.5419}\\
            &F1&\underline{0.4603} &0.2530&0.4229&0.2896&0.4556&0.3794& \textbf{0.6615}\\
    
        \hline
            &ACC & 0.8592 & 0.2952& 0.7828 & 0.5756 & 0.5116 &   \underline{0.8932} &\textbf{0.9380}\\
            BDGP&NMI & 0.7080 & 0.0679& 0.5774 & 0.3124 & 0.4024 &  \underline{0.7411}&\textbf{0.8262} \\
            &F1& 0.8588& 0.2576 & 0.7819 & 0.5816 & 0.5582 &  \underline{0.8924}&\textbf{0.9379} \\
    
       \hline
            
            &ACC& \underline{0.4656} & 0.3181 & 0.4631 & 0.4636 & 0.4402 & 0.2637 &\textbf{0.5416}\\
        Reuters\_dim10&NMI& 0.2212&0.0411  & 0.2261& \underline{0.2487} & 0.1943 &  0.0064& \textbf{0.3313} \\
             &F1& 0.4554&0.2321  & 0.4850 & 0.4886 & \underline{0.5195} & 0.1129&\textbf{0.5636} \\

        \hline
            &ACC&\underline{0.7714}&  0.3839& 0.7143 & 0.4667 & 0.5262 &   0.1524&\textbf{0.7865}\\
            MSRCv1&NMI & \underline{0.6397} & 0.1919 & 0.5657& 0.3291 & 0.4082 &   0.0276&\textbf{0.6884} \\
             &F1& \underline{0.7682} & 0.3851 & 0.7097 & 0.4591 & 0.5524 &   0.0546&\textbf{0.7837} \\
        \hline
            &ACC & 0.2806 & 0.1908 &0.3185 &  \underline{0.3287} & 0.2814 &  0.3285&\textbf{0.8222}\\
      ALOI &NMI & 0.5373 &  0.3459& 0.5683 &  0.5445 & \underline{0.5725} &  0.4839 & \textbf{0.8801}\\
             &F1&0.2513 &0.1994 & 0.3000 & 0.3173 & \underline{0.3370} & 0.3330&\textbf{0.8192} 
            \\
              \hline
        &ACC &\underline{0.4497} &0.3224&0.4412&0.4115&0.3766&0.2497&\textbf{0.5930}\\
        yale\_mtv&NMI & \underline{0.5287}&0.3860&0.5170&0.4768&0.0759&0.3825&\textbf{0.6129} \\
        &F1&\underline{0.4394} &0.3101&0.4244&0.4032&0.4176&0.2030&\textbf{0.6166} \\
    
       \hline
      
    \end{tabular} }
\end{table*}

\section{Experiments}
\label{sec:Experiments}

\begin{table*}[!htbp]
    \centering
    \caption{Introduction to datasets}
    \label{tbl:datasets}
    \resizebox{\textwidth}{!}{
    \begin{tabular}{c|ccccccccccc}
        \hline
        Datasets & 3Sources & BBCsports & Caltech101 & BDGP & Handwritten & Movies & flowers17 & Prokaryoic & yale\_mtv & Reuters\_dim10 & ORL \\
        \hline
        Dim 1 & 3560 & 2582 & 1984 & 1750 & 47 & 1878 & 1360 & 393 & 4096 & 10 & 4096 \\
        Dim 2 & 3631 & 2544 & 512 & 79 & 240 & 1398 & 1360 & 438 & 3304 & 10 & 3304 \\
        Instances & 169 & 282 & 2386 & 2500 & 2000 & 617 & 1360 & 551 & 165 & 18758 & 400 \\
        Classes & 6 & 5 & 20 & 5 & 10 & 17 & 17 & 4 & 15 & 6 & 40 \\
        \hline
    \end{tabular}
    }
\end{table*}

This study conducts sample-level alignment experiments on \textbf{16} widely used multimodal datasets and clustering experiments on \textbf{6} widely used multimodal datasets. The model's advancement is demonstrated through alignment accuracy and clustering results. To validate the model's effectiveness in solving the issues of multimodal incompleteness and misalignment, we benchmark it against six state-of-the-art methods proposed in recent years, using ACC, NMI, and weighted F1-score as performance metrics. Experiments are conducted under high alignment rate (0.7), medium alignment rate (0.5), and low alignment rate (0.3).
Considering that models like PVC\cite{huang2020partially}, MvCLN\cite{yang2021partially}, SMILE\cite{zeng2023semantic}, EGPVC\cite{zhao2023end}, and DGPPVC\cite{zhao2024dynamic} only address misalignment, we first preprocess the data by handling missing values and filling them using Gaussian kernel interpolation to ensure the smoothness of the imputed data before conducting the model experiments.

The experiments are conducted in a Python 3.8 environment, using PyCharm 2022.2.3 as the integrated development environment, with the deep learning framework PyTorch 1.8.2 and CUDA version 11.1. Detailed information about the datasets used in the experiments can be found in Table \ref{tbl:datasets},\ref{tbl:datasets2}. Since the main challenge of multimodal alignment problems arises from misalignment, we set the incompleteness rate to 0.5, which is a typical value for this setting.

\begin{table}[!htbp]
    \centering
    \caption{Introduction to datasets}
    \label{tbl:datasets2}
    \resizebox{\columnwidth}{!}{
    \begin{tabular}{c|ccccc}
        \hline
        Datasets & MSRCv1 & 20NewsGroups & BBC4 & ALOI & Wikipedia-test \\
        \hline
        Dim 1 & 24 & 2000 & 4659 & 77 & 128 \\
        Dim 2 & 576 & 2000 & 4633 & 64 & 10 \\
        Instances & 210 & 500 & 685 & 10800 & 693 \\
        Classes & 7 & 5 & 5 & 100 & 10 \\
        \hline
    \end{tabular}
    }
\end{table}

\subsection{Experimental Results and Analysis}
In this experiment, the best-optimized values for each metric are highlighted in bold, while the second-best values are underlined. \(F1\) refers to the F-score(weighted), and \(acc_s\) represents the sample-level alignment accuracy, defined as follows

\begin{equation}
    acc_s = N_t / N_s ,
\end{equation}

where \(N_t\) represents the number of aligned data pairs, and \(N_s\) represents the total number of data pairs.
Table \ref{tblsa} compares the sample-level alignment performance of DLPMAC with six other models. Since PVC, EGPVC, and DGPPVC use soft alignment (based on probability distributions), we use the clustering ACC value instead of \(acc_s\) for these models.

\begin{itemize}
    \item As shown in Table \ref{tblsa}, our model achieves the best performance across all \textbf{16} datasets, significantly improving the sample-level alignment accuracy, thus validating its effectiveness and generalizability in solving the sample-level alignment problem.

    \item This study is the first to explore the alignment problem from the sampling perspective and proposes a model with high generalizability across multimodal datasets. The model addresses the data alignment task and effectively performs post-alignment fusion clustering.
\end{itemize}

Tables \ref{tbl0.5}, \ref{tbl0.3}, and \ref{tbl0.7} report the experimental results of \textbf{7} different methods on \textbf{6} multimodal datasets, with evaluation metrics including ACC, NMI, and F1. These metrics are used to assess the performance of each method in clustering tasks. From these results, we draw the following conclusions.

\begin{itemize}
    \item Tables \ref{tbl0.5}, \ref{tbl0.3}, and \ref{tbl0.7} present the clustering results of the model using seven different methods on six datasets. Our model achieves the optimal values across all metrics on every dataset, demonstrating its strong performance in addressing the challenges of multimodal incompleteness and misalignment.

    \item On the ALOI dataset, for different alignment rates (0.3, 0.5, 0.7), our model significantly outperforms other models in clustering tasks, proving its effectiveness in handling multimodal clustering problems with multiple data and classes.

    \item Furthermore, our model achieves the best results on all datasets, indicating its ability to perform well on both small datasets with fewer classes and large datasets with more classes. This demonstrates that, to these ends, we propose a dual-learning based penalized multi-align clustering model (DLPMAC). Our model can effectively capture structural information between data and learn the underlying features of the data.
\end{itemize}

\begin{figure}
	\centering
        \includegraphics[scale=0.43]{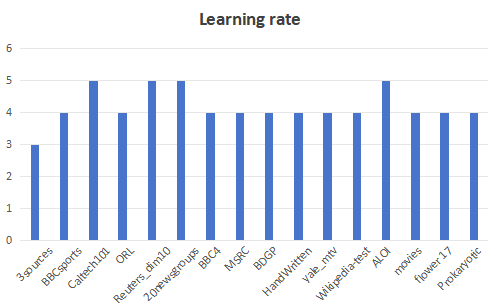}
	\caption{Learning rates of different datasets}
        \Description{A bar chart illustrating the learning rates across different datasets.}
	\label{learn}
\end{figure}

\begin{figure}
	\centering
        \includegraphics[scale=0.09]{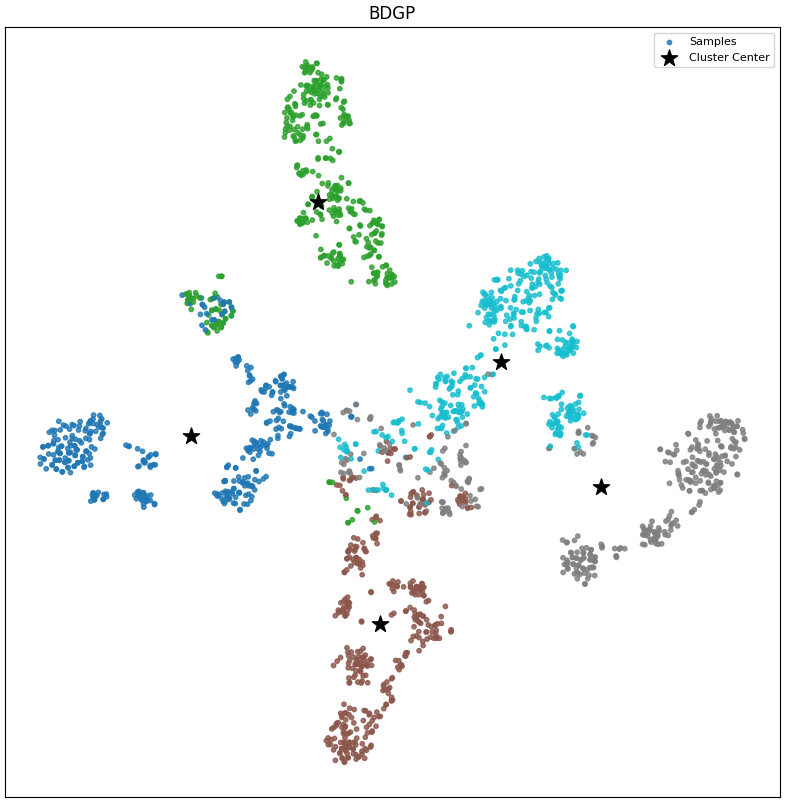}
        \hspace{0.1in}
        \includegraphics[scale=0.09]{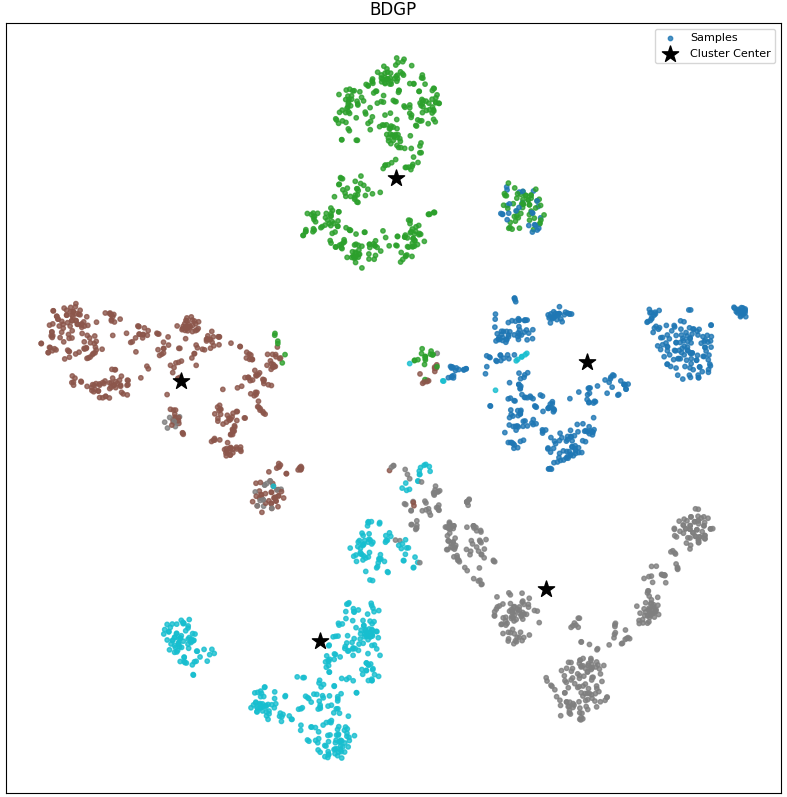}
        \hspace{0.1in}
        \includegraphics[scale=0.09]{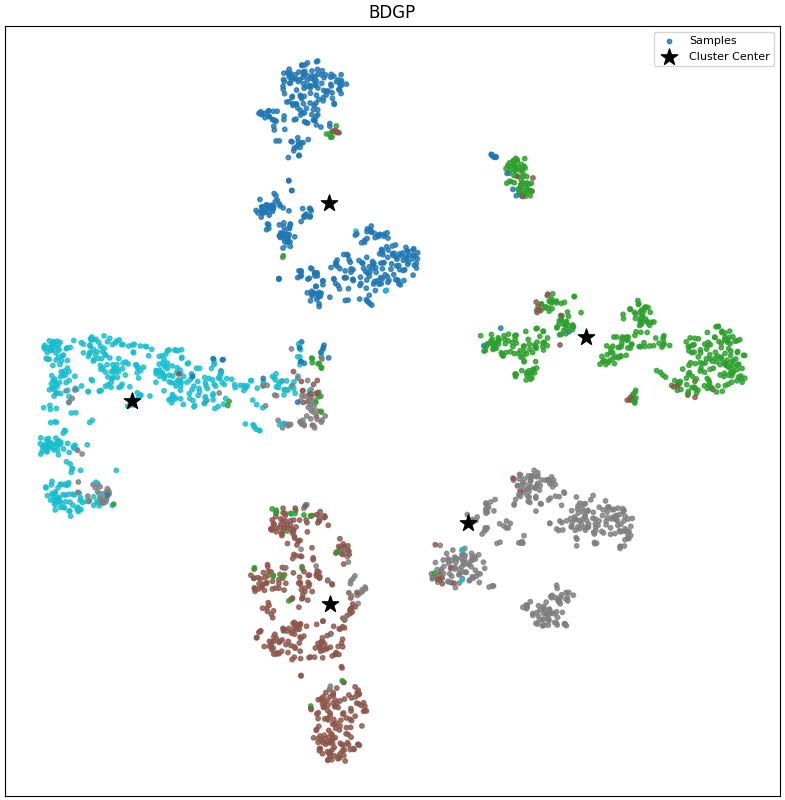}
	\caption{Clustering results of alignment rates (0.3, 0.5, 0.7)}
        \Description{Three clustering plots corresponding to alignment rates 0.3, 0.5, and 0.7, showing colored clusters in 2D space.}
	\label{clu}
\end{figure}

\subsection{Ablation Experiment}

In this section, we validate the effectiveness of our proposed Penalized Multi-Align module through ablation experiments. Specifically, on the one hand, we perform data alignment and fusion using a one-to-one alignment method and discard the unaligned data; on the other hand, we use the Penalized Multi-Align module for data alignment and fusion. We conducted experiments on \textbf{6} datasets using \textbf{3} evaluation metrics. Here, \textbf{O2O} denotes the one-to-one alignment method, and \textbf{M2M} refers to the multi-to-multi alignment method.

\begin{table}
\centering
\caption{Clustering results of ablation experiments}
\label{ab}
\resizebox{\columnwidth}{!}{
\begin{tabular}{c|cc|cc|cc}
\hline
 Methods & O2O & M2M & O2O & M2M&O2O & M2M \\ \hline
 & \multicolumn{2}{c|}{3sources} & \multicolumn{2}{c|}{BDGP} & \multicolumn{2}{c}{Reuters\_dim10}   \\ 
ACC & 0.6204 & \textbf{0.6346} & 0.8819 & \textbf{0.9403} & 0.4199 & \textbf{0.5257} \\ 
NMI & 0.4731 & \textbf{0.5423} & 0.7085 & \textbf{0.8372} & 0.1665 & \textbf{0.3173} \\ 
F1 & 0.6022 & \textbf{0.6568} & 0.8815 & \textbf{0.9407} & 0.4396 & \textbf{0.5531} \\ \hline
 & \multicolumn{2}{c|}{MSRSv1} & \multicolumn{2}{c|}{ALOI} & \multicolumn{2}{c}{yale\_mtv} \\ 
ACC & 0.7403 & \textbf{0.7748} & 0.6673 & \textbf{0.7744} & 0.5141 & \textbf{0.5861} \\ 
NMI & 0.6298 & \textbf{0.6694} & 0.7794 & \textbf{0.8488} & 0.5454 & \textbf{0.6135} \\ 
F1 & 0.7454 & \textbf{0.7744} & 0.6599 & \textbf{0.7730} & 0.5233 & \textbf{0.607} \\ \hline
\end{tabular}}
\end{table}

As shown in Table \ref{ab}, the \textbf{M2M} alignment method improves the model's clustering performance to some extent. Particularly on the larger datasets, Reuters\_dim10 and ALOI, the ACC, NMI, and F1 metrics improved by 10.58\%, 15.08\%, and 11.35\%, and 10.71\%, 6.94\%, and 11.31\%, respectively. This indicates that, in large-scale incomplete and disordered data, data missingness significantly affects subsequent alignment operations, potentially causing many data points to remain unmatched. In this case, using the \textbf{M2M} alignment method allows for better matching of similar data, thereby enhancing clustering performance.

\subsection{Complementary Experiments}
This section presents the clustering results of the model on the BDGP dataset with an alignment rate of 0.3,0.5 and 0.7, as shown in Figure \ref{clu}. From the figure, it is evident that our model performs well in clustering on the BDGP dataset. In addition, we provide the learning rates corresponding to each dataset used in the experiments, as shown in Figure \ref{learn}. Let \(\beta\) represent the vertical axis value, then the learning rate is defined as

\begin{equation}
       Learning\_rate=1 \times 10^{-\beta} .
\end{equation}

\section{Conclusion}
\label{sec:Conclusion}

In this paper, we propose the DLMMAC model to address the issues of incompleteness and misalignment in multimodal data. The model not only effectively solves the sample-level data alignment problem but also improves the alignment at the clustering level. By proposing a Dual-Learning mechanism, we ensure that the model learns the prior knowledge of each modality—specifically, the semantic consistency of the data and the structural similarity of the modalities. Additionally, the design of the Penalized Multi-Align module allows a single sample to form data pairs with different samples from other modalities, improving the alignment accuracy of data pairs and preventing clustering phenomena between data. In the future, we will focus more on semantic structure learning for multimodal data with low alignment rates, further addressing the issue of multimodal incompleteness and misalignment in the absence of prior knowledge.

\begin{acks}
This work was supported by the Science and Technology Project of Liaoning Province (2024JH2/102600027, 2023JH2/101700363), 
the Science and Technology Project of Dalian City (2024JJ12GX025, 2023JJ12SN029, and 2023JJ11CG005), 
and the Hainan Provincial Natural Science Foundation of China under Grant No. 825CXTD608.
\end{acks}

\bibliographystyle{ACM-Reference-Format}
\bibliography{sample-base}

\appendix

\end{document}